\documentclass[acmtog]{acmart}
\acmSubmissionID{1416}

\usepackage{booktabs} 

\usepackage[ruled]{algorithm2e} 

\SetAlFnt{\small}
\SetAlCapFnt{\small}
\SetAlCapNameFnt{\small}
\SetAlCapHSkip{0pt}

\acmJournal{TOG}
\acmVolume{44}
\acmNumber{6}
\acmArticle{243}
\acmYear{2025}
\acmMonth{12}

\setcopyright{cc}

\acmDOI{10.1145/3763302}

\newcommand{\ourmethod}{SS4D}

\begin{document}
\title{{\ourmethod}: Native 4D Generative Model via Structured Spacetime Latents}

\author{Zhibing Li}
\authornote{Equal contribution.}
\orcid{0009-0002-4528-5495}
\affiliation{%
 \institution{The Chinese University of Hong Kong}
 \country{China}}
\affiliation{%
 \institution{Shanghai AI Laboratory}
 \country{China}
}
\email{lz022@ie.cuhk.edu.hk}

\author{Mengchen Zhang}
\authornotemark[1]
\orcid{0009-0004-0141-3939}
\affiliation{%
 \institution{Shanghai AI Laboratory}
 \country{China}
}
\affiliation{%
 \institution{Zhejiang University}
 \country{China}
}
\email{zhangmengchen@zju.edu.cn}

\author{Tong Wu}
\authornote{Corresponding Authors.}
\orcid{0000-0001-5557-0623}
\affiliation{%
\institution{Stanford University}
\country{USA}}
\email{wutong16@stanford.edu}

\author{Jing Tan}
\orcid{0009-0005-8016-915X}
\affiliation{%
 \institution{The Chinese University of Hong Kong}
 \country{China}}
\email{tj023@ie.cuhk.edu.hk}

\author{Jiaqi Wang}
\orcid{0000-0001-6877-5353}
\email{wjqdev@gmail.com}
\affiliation{%
	\institution{Shanghai AI Laboratory}
	\country{China}
}

\author{Dahua Lin}
\authornotemark[2]
\orcid{0000-0002-8865-7896}
\email{dhlin@ie.cuhk.edu.hk}
\affiliation{%
    \institution{The Chinese University of Hong Kong}
	\country{China}
}


\begin{abstract}

We present {\ourmethod}, a native 4D generative model that synthesizes dynamic 3D objects directly from monocular video.
Unlike prior approaches that construct 4D representations by optimizing over 3D or video generative models, we train a generator directly on 4D data, achieving high fidelity, temporal coherence, and structural consistency.
At the core of our method is a compressed set of structured spacetime latents. Specifically,
\textbf{(1)} To address the scarcity of 4D training data, we build on a pre-trained single-image-to-3D model, preserving strong spatial consistency. 
\textbf{(2)} Temporal consistency is enforced by introducing dedicated temporal layers that reason across frames. 
\textbf{(3)} To support efficient training and inference over long video sequences, we compress the latent sequence along the temporal axis using factorized 4D convolutions and temporal downsampling blocks.
In addition, we employ a carefully designed training strategy to enhance robustness against occlusion and motion blur, leading to high-quality generation.
Extensive experiments show that {\ourmethod} produces spatio-temporally consistent 4D objects with superior quality and efficiency, significantly outperforming state-of-the-art methods on both synthetic and real-world datasets.

\end{abstract}

%
%



%
%


\begin{teaserfigure}
  \centering
  \includegraphics[width=\linewidth]{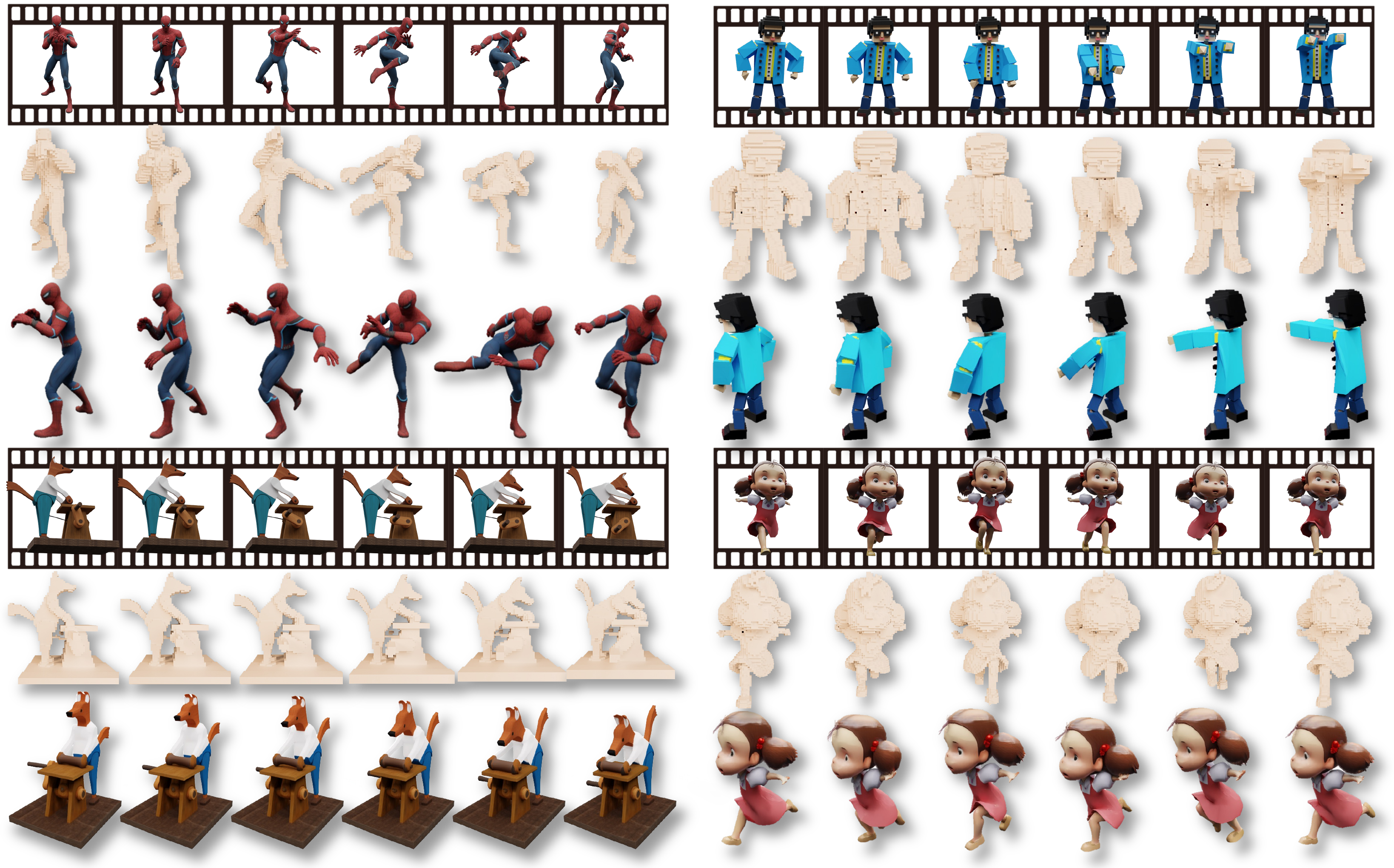}
  \caption{\small {\ourmethod} generates high-quality 4D content in 2 minutes. The monocular input videos, corresponding voxelized structure, and the final 4D content from an alternative viewpoint are presented. For additional dynamic results, please refer to our supplementary demo video. \url{https://lizb6626.github.io/SS4D/}}
  \label{fig:teaser}
\end{teaserfigure}

\maketitle

\section{Introduction}

Bridging the digital and physical worlds stands as one of the most profound frontiers in computer graphics and vision. While static 3D object creation has seen remarkable progress~\cite{zhang2024clay,xiang2024structured,hunyuan3d22025tencent}, the synthesis of dynamic 3D content, also known as 4D, has lagged behind. 4D generation seeks to model not only how objects look but also how they move through time, holding the potential to redefine the fields of filmmaking, gaming, and virtual reality. In this work, we propose a method for efficient and high-quality 4D object generation from monocular videos, enabled by a native 4D latent space that captures both spatial structure and temporal dynamics.

Previous approaches to 4D content creation generally fall into two categories. The first line of works~\cite{jiang2024consistentd,bah20244dfy,zeng2024stag4d,zhang20244diffusion,ling2024align} leverages pre-trained multi-view and video generative models to optimize 4D representations—such as 4D Gaussians and Dy-NeRF—through intricate Score Distillation Sampling (SDS)~\cite{poole2022dreamfusion} and its variants. While conceptually appealing, SDS-based methods often suffer from over-saturation artifacts and require hours of per-instance optimization. Another line of works\cite{yang2024diffusion, sun2024eg4d} refines video models for multi-view video generation and reconstructs 4D objects directly via photogrammetry losses. Though faster, these methods tend to yield coarse or noisy geometry and struggle to maintain spatio-temporal consistency, particularly under highly dynamic motion.

Recent advances in native 3D generative models~\cite{zhao2023michelangelo, zhang2024clay, xiang2024structured, hunyuan3d22025tencent, li2025triposg} trained on large-scale 3D datasets~\cite{objaverse,objaverseXL} have demonstrated notable improvements in both quality and efficiency over SDS-based approaches~\cite{poole2022dreamfusion, lin2023magic3d, chen2023fantasia3d, liang2024luciddreamer, qiu2024richdreamer} and Large Reconstruction Models (LRMs)~\cite{hong2023lrm, li2023instant3d, gslrm2024}. Inspired by this evolutionary trajectory in 3D generation, we extend it into the 4D domain. Our method builds upon TRELLIS~\cite{xiang2024structured}, which encodes 3D objects as sparse voxel grids enriched with attribute channels, offering strong spatial inductive biases through the spatial locality inherent in sparse voxel structures.

We propose \ourmethod, a native 4D generative model capable of synthesizing dynamic 3D objects directly from monocular video input. Our approach extends TRELLIS into the spacetime domain by incorporating time as the fourth dimension in the latent space, while retaining the strong spatial consistency provided by its structured latents representation.
Specifically, we fine-tune TRELLIS's autoencoder and generator with Temporal Layers and carefully designed 4D positional encodings to enforce consistency across frames. To improve long-term sequence generation, we apply a temporal downsampling strategy via dedicated downsampling blocks and factorized 4D convolutions. Furthermore, we introduce a progressive training schedule and a random masking strategy to enhance robustness against occlusions and motion blur, both of which are common challenges in real-world video data.

 
We conduct extensive experiments on synthetic benchmarks, including ObjaverseDy~\cite{xie2024sv4d} and Consistent4D~\cite{jiang2024consistentd}, as well as real-world video sequences from DAVIS~\cite{DAVIS_2019}. Quantitative and qualitative results show that {\ourmethod} achieves state-of-the-art performance, generating high-quality, spatio-temporally consistent 4D objects with remarkable computational and memory efficiency. The generated 4D content and its voxelized structure by {\ourmethod} are visualized in Figure~\ref{fig:teaser}. We believe our model can serve as a foundation for future work in 4D generation and unlock new possibilities in dynamic content creation.

\begin{figure*}[t]
	\centering
	\includegraphics[width=1.0\linewidth]{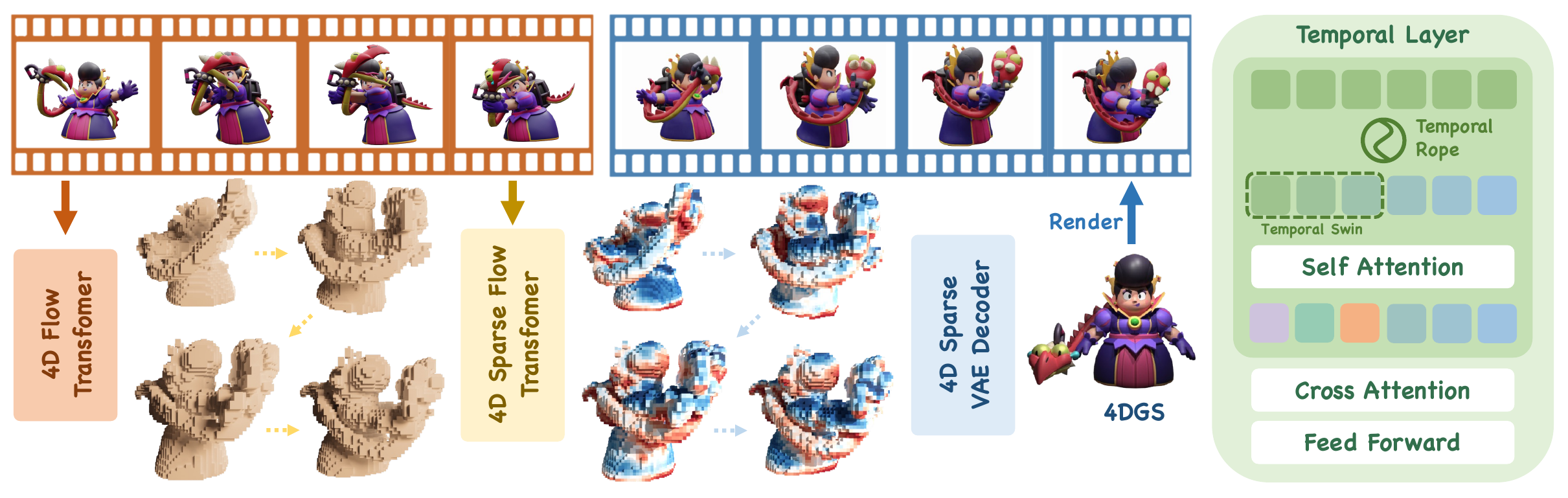}
    \caption{
    \textbf{\small {\ourmethod} Overview.} \textbf{Left}: {\ourmethod} pipeline. Our method takes a monocular video as input. It extracts a coarse voxel-based structure using a 4D Flow Transformer, then generates spacetime latents through a 4D Sparse Flow Transformer, both incorporating a Temporal Layer to capture temporal consistency. These latents are subsequently decoded into a sequence of 3D Gaussians, forming the final 4D content. \textbf{Right}: Temporal Layer. It combines temporal self-attention with shifted windows and hybrid 1D Rotary Position Embeddings to efficiently model dynamic 4D content and ensure consistency across frames.}
	\label{fig:overview}
\end{figure*}

\section{Related Work}

\paragraph{SDS-based 4D Generation}
A prominent line of research in 4D generation~\cite{jiang2024consistentd,bah20244dfy,ling2024align,zheng2024unified,ren2023dreamgaussian4d,yu20244real,zeng2024stag4d} leverages Score Distillation Sampling (SDS) with off-the-shelf pretrained models. These methods distill different types of priors: appearance priors: appearance priors~\cite{ling2024align} from text-to-image diffusion models, 3D priors~\cite{zeng2024stag4d,ren2023dreamgaussian4d} from 3D-aware and multi-view diffusion models, and the motion priors~\cite{ling2024align} from video diffusion models. 
Consistent4D~\cite{jiang2024consistentd} optimizes a dynamic NeRF conditioned via SDS loss derived from 3D-aware image diffusion.
4D-fy~\cite{bah20244dfy} sequentially combines gradients from three distinct pretrained diffusion models to harness the strengths of each modality, thereby enhancing the quality of 4D generation.
While SDS-based approaches effectively utilize diffusion priors to generate dynamic motion, they remain computationally expensive, often requiring several hours of optimization per object, and are prone to over-saturation artifacts.

\paragraph{Feed-forward 4D Generation}
Another line of work in 4D generation~\cite{yang2024diffusion,sun2024eg4d,xie2024sv4d,yao2024sv4d2,wu2024cat4d} adopts feed-forward pipelines that avoid iterative optimization with SDS gradients. Diffusion$^2$~\cite{yang2024diffusion} and EG4D~\cite{sun2024eg4d} leverage off-the-shelf video diffusion and multi-view diffusion to generate images at different viewpoints and timesteps using training-free techniques, followed by 4D reconstruction from the synthesized images. 
Other approaches~\cite{xie2024sv4d,yao2024sv4d2,wu2024cat4d} train multi-view multi-time diffusion with 4D datasets~\cite{xie2024sv4d} or through a combination of multi-view and video datasets.
L4GM~\cite{ren2024l4gm} extends LGM~\cite{tang2024lgm} to generate 3D Gaussians at each frame.
However, most of these feed-forward methods rely on minimizing volumetric rendering losses on RGB images, which often leads to coarse or noisy geometry in the resulting 4D representations.

More recent methods aim to improve temporal consistency and geometric fidelity. V2M4~\cite{chen2025v2m4} first generates 3D meshes for individual frames, then enforces consistency through mesh registration and texture optimization. AnimateAnyMesh~\cite{wu2025animateanymesh} and Puppeteer~\cite{song2025puppeteer} animate static 3D meshes using text or video prompts. The most relevant work, GVFDiffusion~\cite{zhang2025gaussian} encodes canonical Gaussians and their temporal displacements into a latent space using a 4DMesh-to-GS Variation Field VAE. A Gaussian Variation Field diffusion model is then trained on this latent space to generate 4D content. However, the canonical Gaussians are conditioned only on the first frame, the model may fail to match the appearance in subsequent frames, resulting in suboptimal performance.

Our method directly synthesizes 4D objects using a structured spacetime latent representation. By explicitly generating surfaces with geometric supervision, our framework produces more accurate 4D geometry that remains consistent across large viewpoint changes and aligns with the underlying motion.



\section{Background: Structured 3D Latents}

Our approach is built upon TRELLIS~\cite{xiang2024structured}, a model for generating 3D assets from a single image. TRELLIS introduces structured latents that effectively capture both the geometry and appearance of 3D objects.
TRELLIS first transforms a 3D asset into a voxelized feature representation, defined as:
\begin{equation}
    f = \{(f_i,p_i)\}^L_{i=1}, f_i\in \mathbb{R}^C, p_i\in \{0,1,...,N-1\}^3,
\end{equation}
where $p_i$ denotes the coordinate of an activated voxel in a 3D grid, approximating the coarse surface of the 3D asset. The feature $f_i$ associated with each voxel is computed by aggregating DINOv2~\cite{oquab2023dinov2} features extracted from multi-view renderings. $N$ denotes the spatial resolution of the voxel grid, and $L$ is the total number of activated voxels.

TRELLIS encodes the voxelized feature $f$ into structured latents $z=\mathcal{E}(f)=\{(z_i,p_i)\}^L_{i=1}, z_i\in \mathbb{R}^D$, using 
a 3D Variational Auto-Encoder (VAE)~\cite{vae}.
These latents can then be decoded into various 3D representations, including NeRF~\cite{mildenhall2020nerf}, 3D Gaussians~\cite{kerbl3Dgaussians}, and meshes.

The generation of structured latents is decomposed into two stages: (1) generating the sparse structure $\{p_i\}^L_{i=1}$, followed by (2) generating the latent features $\{z_i\}^L_{i=1}$. Both stages employ generative models ($\mathcal{G}_S$ for structure and  $\mathcal{G}_L$ for latents) that leverage a Transformer-based Diffusion model (DiT)~\cite{dit} architecture and the Conditional Flow Matching (CFM) objective~\cite{lipman2024flow}.

\vspace{-5pt}
\section{Method}

Given a monocular video of a dynamic object $\mathcal{I}=\{I_t\}^T_{t=1}$, where $T$ is number of frames and each $I_t \in \mathbb{R}^{3\times H \times W} $ represents an RGB image, our goal is to produce a high-quality 4D representation of the object. To achieve this, we extend structured 3D latents into structured spacetime latents $Z=\{z_t\}^T_{t=1}$. 
Through this extension, our model inherits spatial consistency from the pre-trained 3D backbones, alleviating the challenge posed by the scarcity of 4D data.

An overview of our model is illustrated in Figure~\ref{fig:overview}. We begin by generating a coarse voxel-based structure $P=\{p_t\}^T_{t=1}$ from the input video. We then generate sparse structured spacetime latents $Z$, which are subsequently decoded into a set of 3D Gaussians $\mathcal{K}=\{K_t\}^T_{t=1}$, representing the object at each time step.

\subsection{Temporal Alignment}
\label{sec:temporal}

A key insight in training a native 4D generator is to leverage a pre-trained 3D generation model. However, extending a 3D latent space to a structured spacetime representation is non-trivial. While such models can produce high-quality 3D assets for individual frames, directly applying them to dynamic videos leads to noticeable temporal inconsistencies, 
as they lack temporal awareness.

To address this, we extend the spatial self-attention layers of the pre-trained model into temporal self-attention layers. Specifically, we rearrange the temporal axis into the length dimension to perform attention, followed by reshaping to restore the original structure.
The process, using \texttt{einops}~\cite{rogozhnikov2022einops} notation, is as follows:
\begin{equation}
\begin{aligned}
    z &= \text{rearrange}(z, (\mathrm{B}\,\mathrm{T}) \ \mathrm{M} \ \mathrm{C} \rightarrow \mathrm{B}\ (\mathrm{T} \, \mathrm{M}) \ \mathrm{C} ), \\
    z &= \text{TemporalAttn}(z), \\
    z & = \text{rearrange}(z, \mathrm{B}\ (\mathrm{T} \, \mathrm{M}) \ \mathrm{C} \rightarrow (\mathrm{B}\,\mathrm{T}) \ \mathrm{M} \ \mathrm{C}),
\end{aligned}
\end{equation}
where \( \mathrm{B} \) represents the batch size, \( \mathrm{T} \) is the number of temporal frames, \( \mathrm{M} \) is the attention length, and \( \mathrm{C} \) is the feature dimension. 

Given the local nature of temporal sequences and the quadratic complexity of full attention with increasing frame counts, we adopt shifted window attention~\cite{liu2021swin} in temporal layers to improve efficiency. The temporal layers alternate between non-shifted and shifted window configurations, enabling both local context aggregation and global information exchange.

An essential component of temporal attention is positional embedding. TRELLIS uses absolute position embedding~\cite{vaswani2017attention} to encode 3D spatial coordinates. We retain this design to preserve spatial reasoning capabilities and further incorporate 1D Rotary Positional  Embedding (RoPE)~\cite{su2024roformer} along the temporal axis to capture dynamic relationships between adjacent latents. This hybrid 4D positional encoding allows our model to generalize across varying sequence durations, despite being trained on a fixed number of frames. The complete design of the temporal layer is shown on the right side of Figure~\ref{fig:overview}.

We find that temporal alignment of the diffusion model alone is insufficient. Since the VAE is trained exclusively on static 3D objects, it introduces flickering artifacts when encoding and decoding temporally coherent 4D features. To address this, we apply the same temporal alignment strategy to the VAE, enabling it to produce more coherent 4D reconstructions. As demonstrated in Section~\ref{sec:ablation}, this alignment significantly reduces jittering in the reconstructed Gaussians and is critical for achieving high-quality results.

\subsection{Long-Term Generation}
\label{sec:compression}
 
\begin{figure}[t]
	\centering
	\includegraphics[width=0.95\linewidth]{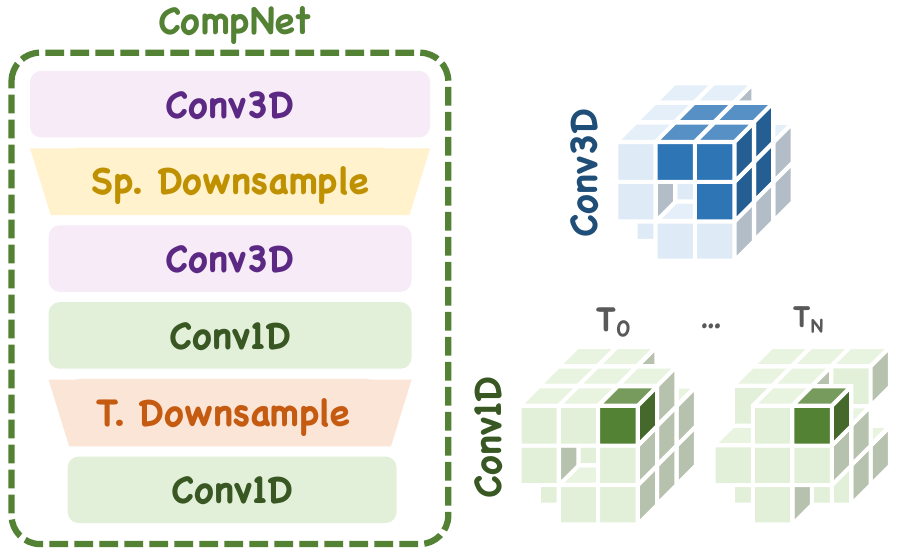}
        \caption{\small \textbf{Implementation of our 4D Compression Strategy.} The strategy improve efficiency by compressing 4D representations for long-term sequence generation in our 4D Sparse Flow Transformer.}
	\label{fig:compression}
\end{figure}

Another major challenge in 4D generation is producing continuous 4D sequences from long videos. Previous works~\cite{jiang2024consistentd,zeng2024stag4d, ren2023dreamgaussian4d} often encounter issues such as quality degradation and inconsistency when generating long sequences, with optimization times increasing drastically. Given limited computational resources, generating spatiotemporally consistent 4D sequences remains a significant challenge. 
To tackle this issue, {\ourmethod} incorporates the 1D Rotary Position Embedding (RoPE)~\cite{su2024roformer}, which enhances the model's ability to extrapolate and generate 4D objects that extend beyond the training dataset. More importantly, considering the significant overlap of information between frames, we improve efficiency by compressing 4D representations of long sequences.


In the 4D Sparse Flow Transformer stage, we generate structured spacetime latents \( \{z_i\}^L_{i=1} \) based on coarse structure \( \{p_i\}^L_{i=1} \). As shown in Figure~\ref{fig:compression}, we explore a 4D compression and convolutional network \textit{CompNet}, which is specifically designed to compress and exchange information across frames. The network captures long-range dependencies while preserving fine-grained spatial details and temporal consistency, ultimately enhancing the quality of the generated 4D sequences.

Specifically, given the input structured spacetime latents \( z \in \mathbb{R}^{L \times D} \), we first apply spatial 3D convolution and compression to each individual frame. A sparse 3D convolution block~\cite{3dsparseconv, spconv2022} with space downsampling is employed to aggregate the latents within a local region of size \( 2^3 \). Then, we apply sparse 1D convolutions~\cite{spconv2022} to facilitate information exchange across frames. Following this, a temporal downsampling block is introduced to compress the latent features by packing two active voxels from the same \( xyz \) position across frames. This module effectively compresses the frame length and reduces the computational load in subsequent Transformer modules.





After the above compression, the timesteps are integrated through AdaLN layers, and video conditions are injected via multiple Transformer blocks, each containing a Temporal Layer. The subsequently compressed latent \( z \) is decompressed in the reverse order of CompNet. Meanwhile, during the decompression process, the convolutional upsampling block is connected to the downsampling block from the compression process through skip connections, facilitating the flow of spatial information. Finally, we restore the latent \( z \) to its original structure.


\subsection{Training \ourmethod}
\label{sec:training}

\paragraph{Data Curation} We curate a dataset of 16,000 animated 3D objects from Objaverse~\cite{objaverse} and ObjaverseXL~\cite{objaverseXL}, filtering out samples with low visual quality or minimal motion. During feature aggregation to activate voxels, we consider only the rendered views in which a voxel is visible, rather than averaging across all views. Voxels that remain invisible in all views are discarded. This curation process not only reduces the length of the spacetime latents, enabling more efficient training, but also minimizes feature noise, leading to improved reconstruction quality.

\paragraph{Progressive Learning} Animated objects from the Internet have different frame lengths. And directly training on long sequence causes slow to convergence. To fully utilize data and save training cost, we employ a progressive learning strategy. That is, we first train on short animations to learn dynamic relationship among spacetime latents. Finally, we select a subset of high-quality and long duration 4D data to finetune our model to cope with long sequences.

\paragraph{Random Masking Augmentation} Real-world objects often appear in complex environments, where they may be partially occluded by surrounding elements or exhibit rapid motion that results in motion blur—such as during dancing or fast movement. To improve robustness to these real-world challenges, we introduce a simple yet effective data augmentation strategy: randomly applying black masks to the conditioning video frames. This augmentation simulates common artifacts like occlusion and motion blur. Experimental results demonstrate that this approach mitigates the impact of such challenges on generation quality.

\section{Experiments}

\begin{figure*}[t]
	\centering
	\includegraphics[width=1.0\linewidth]{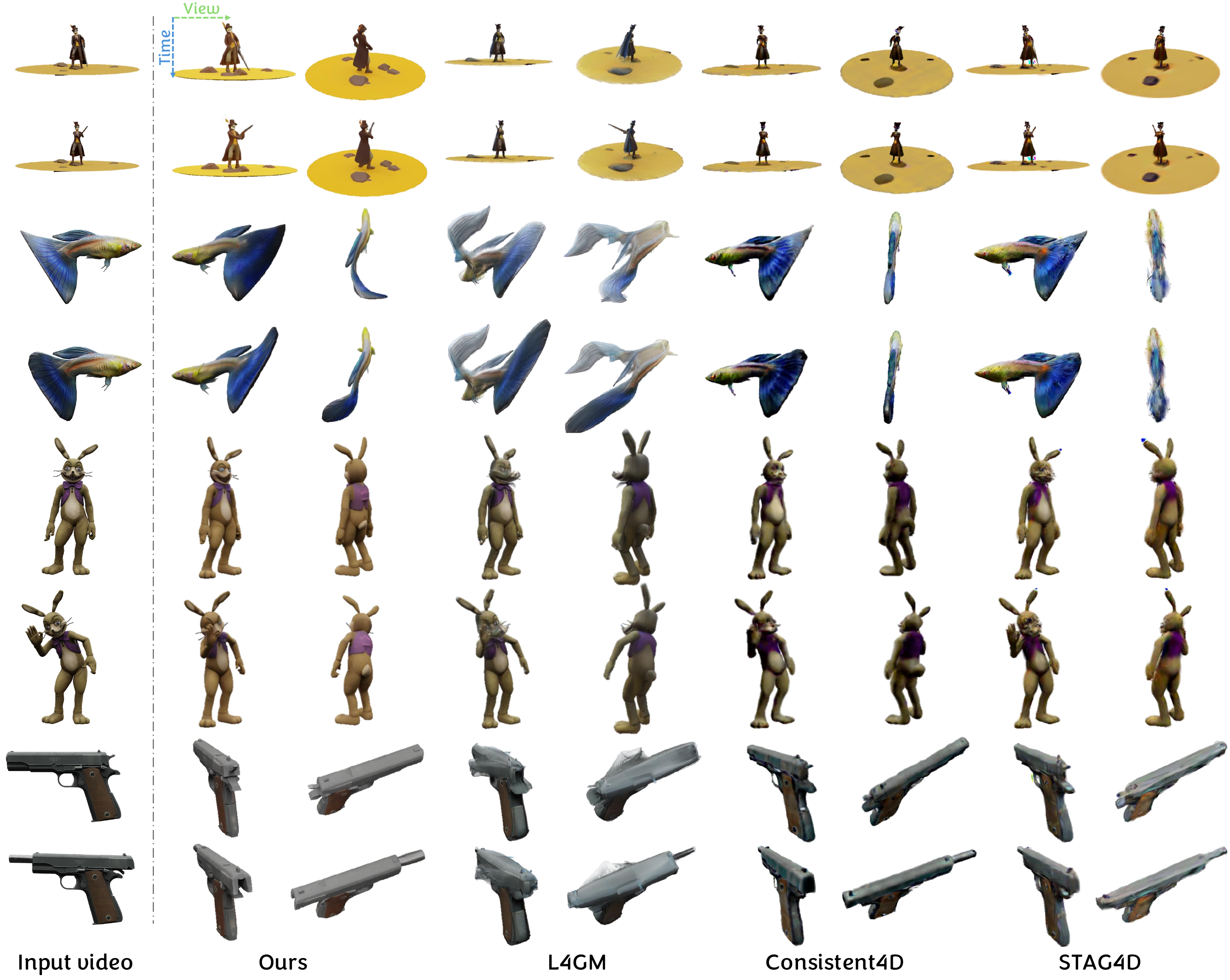}
    \caption{\small\textbf{Qualitative Comparisons on Synthetic Data.} We show two frames from the conditioning video alongside two corresponding novel view renderings. Compared to the baselines, our method generates more accurate geometry and detailed textures that remain consistent over time. }
	\label{fig:synthesis}
\end{figure*}

\begin{figure*}[t]
	\centering
	\includegraphics[width=1.0\linewidth]{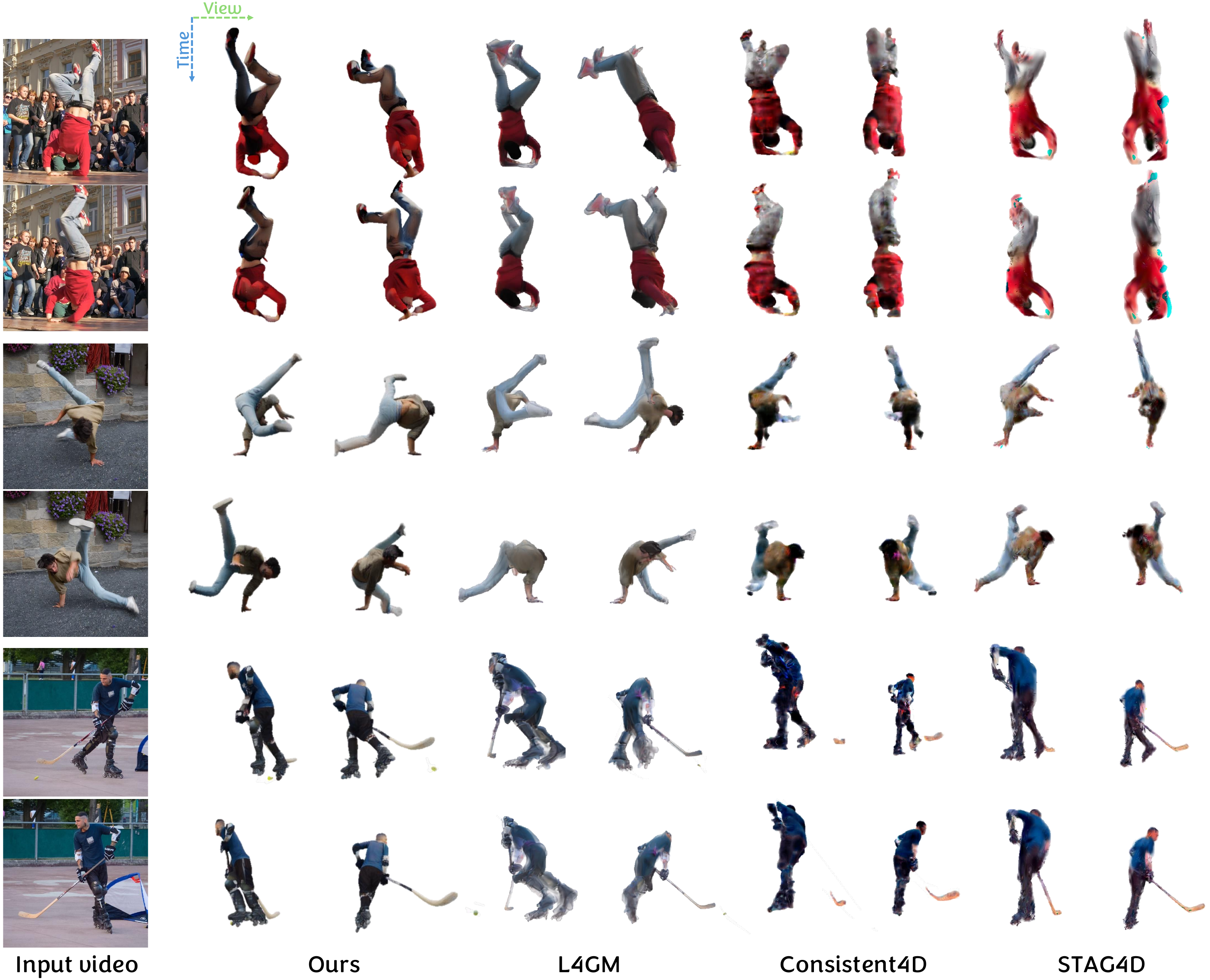}
    \caption{\textbf{Qualitative Comparisons} on real-world videos from \textbf{DAVIS}. For each method, we show two frames from the conditioning video under two novel viewpoints in a $2\times 2$ grid. Compared to the baseline methods, our model remains robust to real-world cases with more complex and high-speed motion.}
	\label{fig:davis}
\end{figure*}

\paragraph{Implementation Details} We rendered our training data using Blender\footnote{https://www.blender.org/} with the Cycles engine. For each animation, we use the first 36 frames if the sequence exceeds that length. 
Following TRELLIS, our pipeline consists of four models: the 4D Structure VAE, 4D Flow Transformer, 4D Sparse VAE, and 4D Sparse Flow Transformer. Among these, the 4D Structure VAE is frozen during training, as it achieves near-perfect reconstruction of the 4D structure sequence. The remaining three models are fine-tuned.
Training is performed on 8×A800 GPUs using the AdamW optimizer with a learning rate of $1\text{e}{-4}$ and FP16 mixed precision. The batch size is set to 2 per GPU for the generator and 1 per GPU for the VAE. Training takes approximately 7 to 8 days in total. For the progressive learning strategy, we start with 8-frame sequences and gradually increase the sequence length to 16 and then 32 frames as training progresses.

\paragraph{Datasets} We evaluate our method on two synthetic datasets: ObjaverseDy~\cite{xie2024sv4d} and Consistent4D~\cite{jiang2024consistentd}. For each object in these datasets, we uniformly sample 32 camera viewpoints and render up to 32 frames per object at a resolution of $512 \times 512$. Additionally, we render a front-view video per object to serve as input across all methods. To assess robustness and generalization to real-world scenarios, we also incorporate video sequences from DAVIS~\cite{DAVIS_2019} for test.

\paragraph{Evaluation Metrics} To evaluate image quality, we compute Peak Signal-to-Noise Ratio (PSNR) and Structural Similarity Index Measure (SSIM)~\cite{wang2004image} between generated novel views and corresponding ground truth. We also report LPIPS~\cite{zhang2018unreasonable} and CLIP Similarity (CLIP-S) to assess perceptual and semantic similarity. Since our method focuses on object-centric generation, we crop all images using the ground truth bounding box to mitigate the influence of background ratio on metrics. 
For evaluating spacetime consistency of generated 4D data, we compute Frechet Video Distance (FVD)~\cite{unterthiner2018towards} on novel view videos.

\paragraph{Baselines} We compare our results against several state-of-the-art baselines, including three SDS-based methods and one feed-forward method: DreamGaussian4D (DG4D)~\cite{ren2023dreamgaussian4d}, Consistent4D~\cite{jiang2024consistentd}, STAG4D~\cite{zeng2024stag4d} and L4GM~\cite{ren2024l4gm}. 
DG4D and STAG4D both train deformable 3D Gaussians using SDS losses from multi-view diffusion models. Consistent4D employs a cascaded DyNeRF architecture optimized with SDS losses from an image-to-image diffusion model.
L4GM trains a large reconstruction model that directly generates 3D Gaussians for each frame in the conditioning video.

\subsection{Quantitative Comparison} 

We present quantitative comparisons on the ObjaverseDy and Consistent4D datasets in Table~\ref{tab:objaversedy} and Table~\ref{tab:consistent4d}, respectively. Our method consistently outperforms all baselines across all evaluation metrics by a large margin. The results demonstrate the superior visual fidelity and quality of our method. In particular, our method achieves substantially lower FVD scores, highlighting improved spatio-temporal consistency in the generated 4D assets.
Additionally, we present an inference time comparison in Table~\ref{tab:time_comp}. Our method achieves significantly faster inference compared to optimization-based approaches while being slower than L4GM.

\begin{table}[t]
    \centering
    \small
\caption{\textbf{Quantitative Comparisons} on the synthetic dataset \textbf{ObjaverseDy}. \textbf{Bold} denotes the best result, and \underline{underline} the second-best.}
    \vspace{-5pt}
    \begin{tabular}{l|ccccc}
        \toprule
         &  LPIPS$\downarrow$ & CLIP-S$\uparrow$ & PSNR$\uparrow$ & SSIM$\uparrow$ & {FVD}$\downarrow$ \\ 
         \midrule
        DG4D & 0.231 & 0.869 & 13.72 & 0.784 & 693 \\
        Consistent4D & \underline{0.189} & 0.884 & 15.63 & 0.817 & \underline{640} \\
        STAG4D & 0.202 & 0.871 & \underline{15.96} & \underline{0.821} & 716 \\
        L4GM & 0.243 & \underline{0.887} & 14.48 & 0.782 & 686 \\
        \midrule
        Ours & \textbf{0.150} & \textbf{0.932} & \textbf{18.09} & \textbf{0.842} & \textbf{465} \\
        \bottomrule
    \end{tabular}
    \label{tab:objaversedy}
    \vspace{-5pt}
\end{table}

\begin{table}[t]
    \centering
    \small
    \caption{\textbf{Quantitative Comparisons} on the synthetic dataset \textbf{Consistent4D}. \textbf{Bold} denotes the best result, and \underline{underline} the second-best.}
    \vspace{-5pt}
    \begin{tabular}{l|ccccc}
        \toprule
         &  LPIPS$\downarrow$ & CLIP-S$\uparrow$ & PSNR$\uparrow$ & SSIM$\uparrow$ & {FVD}$\downarrow$ \\ 
         \midrule
        DG4D & 0.215 & 0.896 & 15.47 & 0.785 & 641 \\
        Consistent4D & 0.228 & 0.886 & 13.66 & 0.768 & 674 \\
        STAG4D & \underline{0.204} & 0.890 & \underline{16.89} & \underline{0.821} & 740 \\
        L4GM & 0.219 & \underline{0.908} & 15.69 & 0.803 & \underline{596} \\
        \midrule
        Ours & \textbf{0.149} & \textbf{0.947} & \textbf{18.90} & \textbf{0.843} & \textbf{455} \\
        \bottomrule
    \end{tabular}
    \label{tab:consistent4d}
    \vspace{-5pt}
\end{table}

\begin{table}[t]
    \centering
    \small
    \caption{\small\textbf{User Study Results} on real-world videos from \textbf{DAVIS}, evaluating Geometry Quality, Texture Quality, and Motion Coherence of 4D generations. \textbf{Bold} denotes the best result, and \underline{underline} the second-best.}
    \vspace{-5pt}
    \setlength{\tabcolsep}{2pt}
    \scalebox{0.9}{\begin{tabular}{l|ccc}
        \toprule
         &  Geometry Quality $\uparrow$ & Texture Quality$\uparrow$ & Motion Coherence$\uparrow$ \\ 
         \midrule
        DG4D & 2.870& \underline{3.077} & 2.967 \\
        Consistent4D & 2.703 & 2.837 & 2.650  \\
        STAG4D & 1.913 & 1.627 & 1.807 \\
        L4GM & \underline{3.017} & {3.047} & \underline{3.050} \\
        \midrule
        Ours & \textbf{4.497} & \textbf{4.413} & \textbf{4.527} \\
        \bottomrule
    \end{tabular}}
    \vspace{-5pt}
    \label{tab:user_study}
\end{table}

\begin{table}[t]
    \centering
    \small
    \caption{\textbf{Inference Time Comparison.}}
    \vspace{-5pt}
    \begin{tabular}{l|ccccc}
        \toprule
         Method & DG4D & Consistent4D & STAG4D & L4GM & Ours  \\
         \midrule
        Time$\downarrow$ & 15 min & 1.5 hr & 1 hr & \textbf{3.5s} & \underline{2 min} \\
        \bottomrule
    \end{tabular}
    \vspace{-5pt}
    \label{tab:time_comp}
\end{table}

\subsection{Qualitative Comparison}

We present visual comparisons on synthetic data in Figure~\ref{fig:synthesis} and real-world data in Figure~\ref{fig:davis}. L4GM performs well on front views but exhibits noticeable distorted geometry when viewed from large elevation angles. Consistent4D tends to produce over-saturated textures, a common artifact associated with SDS losses, while STAG4D often generates noisy and inconsistent textures. 

In contrast, our method reconstructs accurate and clean geometry, even under complex and high-speed motions (e.g., breakdancing), while maintaining detailed and temporally consistent textures. Furthermore, SDS-based methods typically require several hours to process a single dynamic object, whereas our method completes the same task in approximately 3 minutes.

To further evaluate our performance on real-world data, we conduct a user study comparing our method with the baselines. Following ~\cite{yao2024sv4d2}, we select 14 videos from the DAVIS dataset with relatively stable camera motion. We ask 25 users with background in 3D and video generation to evaluate the 4D generation quality in three aspects: geometry quality, texture quality, and motion coherence. Geometry quality refers to the plausibility of the 3D geometry, whether the generated 3D geometry aligns with the motion. Texture Quality refers to the alignment of texture with the condition video. Motion Coherence refers to the temporal consistency and stability of the generated motion. We employed the Average User Ranking (AUR) metric to evaluate model performance. The users were asked to rate the 4D generations on a scale of 1 to 5, and we report the average user rating in Table~\ref{tab:user_study}. Our model outperformed the baselines in all three aspects with a large margin. STAG4D generally receives the weakest rating for its blurry generations and over-saturated texture. The rest of the methods receive similar scores, with L4GM slightly better in shape quality and motion coherence, and DG4D better in texture quality.
\subsection{Ablation Studies}
\label{sec:ablation}

\begin{figure}[t]
	\centering
	\includegraphics[width=1.0\linewidth]{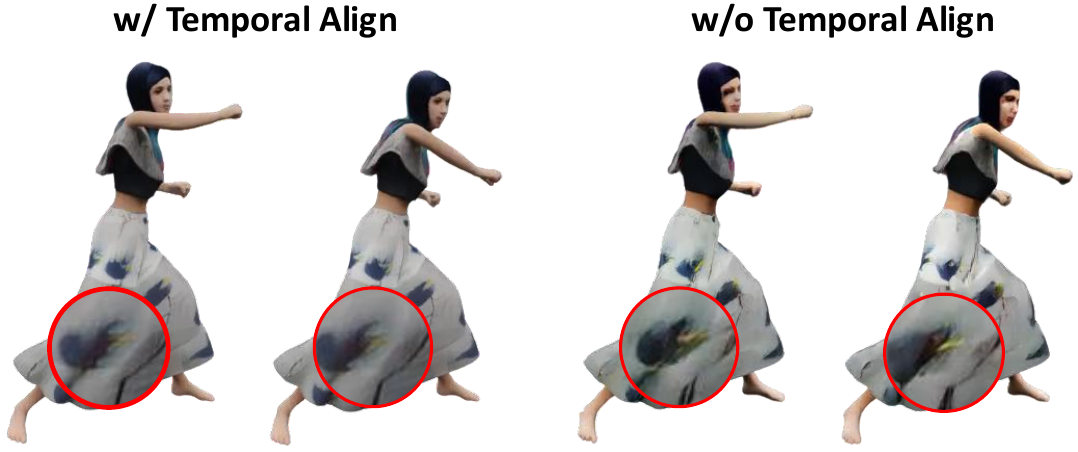}
	\caption{\small
	\textbf{Temporal Alignment in VAE Reconstruction.} Temporal alignment reduces flickering and improves consistency across frames.
	}
	\label{fig:ablation_vae}
\end{figure}

\begin{table}[t]
    \centering
    \small
    \caption{\textbf{Ablation of Temporal Alignment in VAE reconstruction.}}
    \begin{tabular}{l|ccc}
        \toprule
         &  PSNR$\uparrow$ & Flickering$\downarrow$ & FVD$\downarrow$ \\
         \midrule
        w/o Temporal Align & 27.11 & 2.99 & 403.88 \\
        w. Temporal Align & 30.58 & 2.22 & 157.16 \\
        \bottomrule
    \end{tabular}
    \label{tab:ablation_vae}
\end{table}

\paragraph{Temporal Alignment} We ablate the effectiveness of temporal alignment in the VAE reconstruction task. As shown in Figure~\ref{fig:ablation_vae}, without temporal alignment, the reconstructed sequence exhibits noticeable temporal flickering (e.g., the pattern on the dress), whereas incorporating temporal alignment leads to more coherent and stable textures across frames. Following the approach in ~\cite{yang2024cogvideox}, we evaluate two variants using PSNR, FVD and flickering metric that quantifies video flickers by calculating L1 difference between adjacent frames. As reported in Table~\ref{tab:ablation_vae}, temporal alignment significantly reduces flickering and improves the quality of reconstructed 4D assets.

\begin{figure}[t]
	\centering
	\includegraphics[width=1.0\linewidth]{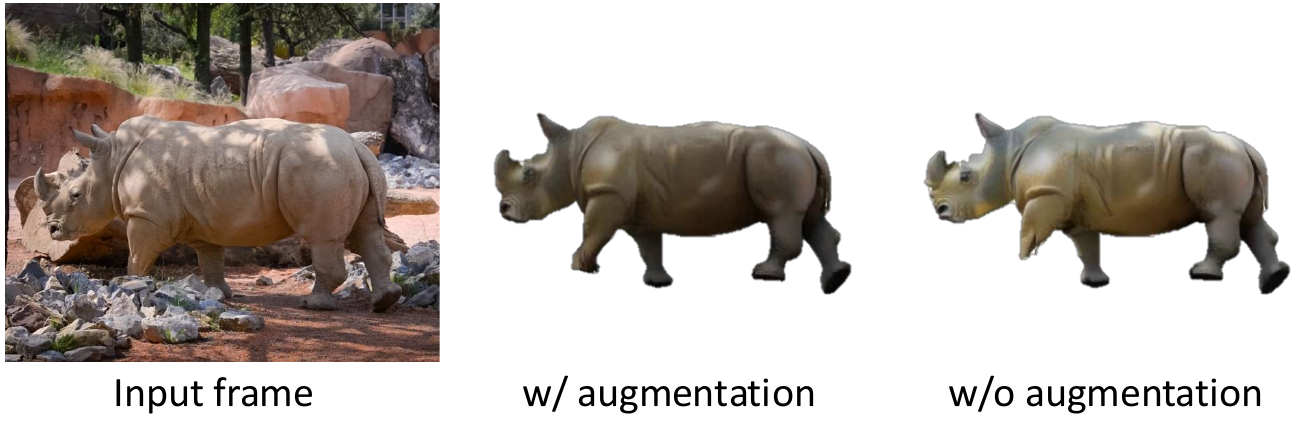}
        \setlength{\abovecaptionskip}{-2mm}
	\caption{\small
	\textbf{Masking Augmentation} enhances the model's robustness to occlusions and improves overall geometry quality.
	}
	\label{fig:ablation_mask}
\end{figure}

\paragraph{Masking Augmentation} As illustrated in Figure~\ref{fig:ablation_mask}, the front leg of the rhino is occluded by rocks and the model without masking augmentation fails to reconstruct the occluded region, resulting in an incomplete front leg. In contrast, applying masking augmentation enables the model to infer and generate the complete shape.

\begin{table}[h]
    \centering
    \small
    \caption{\textbf{Ablation of Feature Aggregation Methods.}}
    \vspace{-8pt}
    \begin{tabular}{l|ccc}
        \toprule
         Aggregation Method &  PSNR$\uparrow$ & Average Length$\downarrow$ & Encode Speed$\uparrow$ \\
         \midrule
        Mean & \textbf{31.63} & 6605 & 68.4 \\
        Visible & 31.07 & \textbf{5261} & \textbf{76.3} \\
        \bottomrule
    \end{tabular}
    \vspace{-10pt}
    \label{tab:ablation_visible}
\end{table}

\paragraph{Visible Feature Aggregation} Averaging all features (as done in TRELLIS) may introduce noise, since occluded voxels can be affected by unrelated views. A rigorous ablation study would require retraining from scratch, which exceeds our  computational resources. Instead, we fine-tune 4D Sparse VAE for 5k steps using visibility-aware aggregation and evaluate reconstruction quality on the ObjaverseDy dataset. As shown in Table~\ref{tab:ablation_visible}, our method achieves comparable reconstruction quality while reducing sequence length and increasing encoding speed (objects/sec), resulting in more efficient training and inference.

\section{Conclusions}

\begin{figure}
    \centering
    \includegraphics[width=0.9\linewidth]{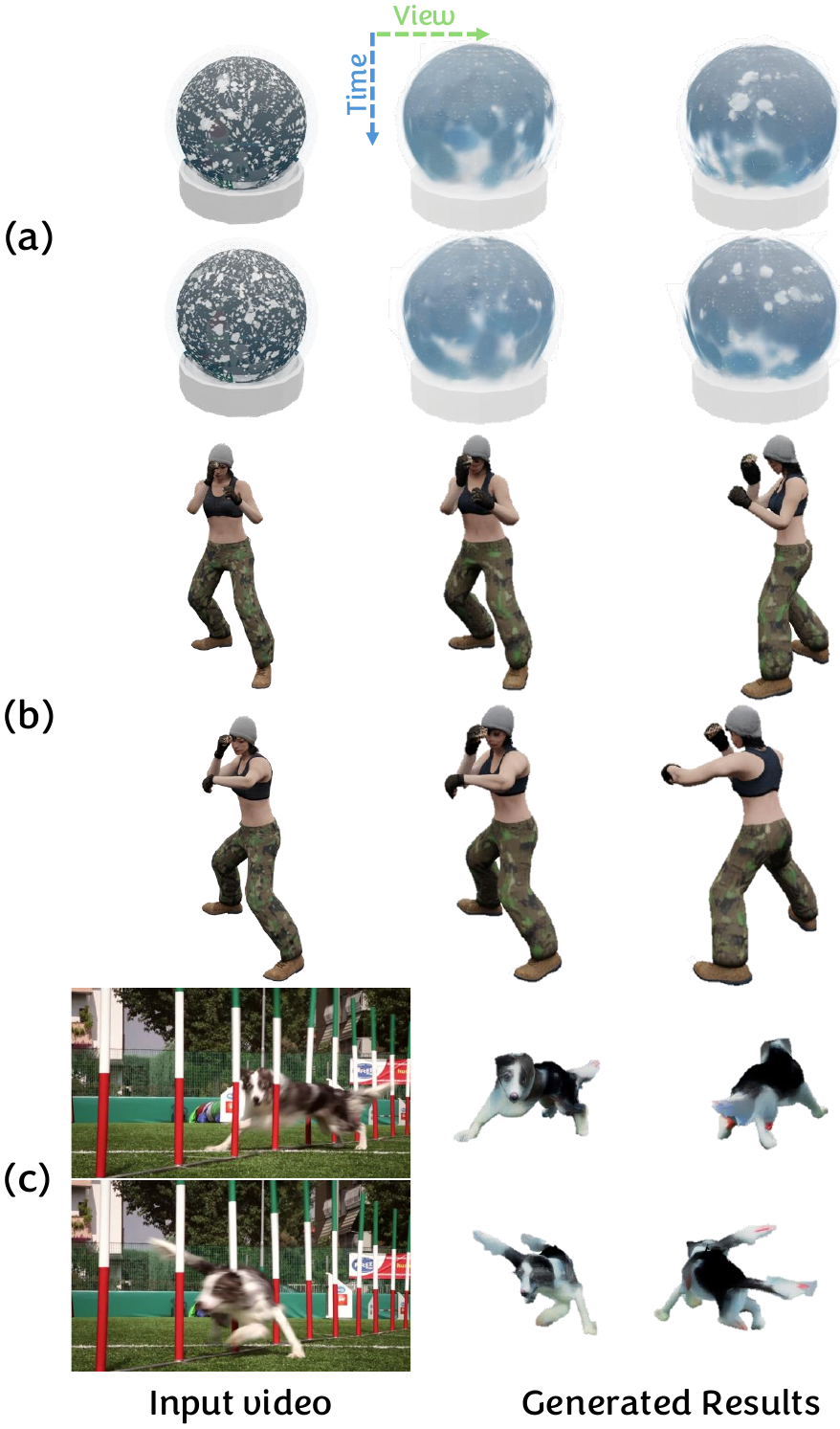}
    \caption{\textbf{Failure Cases.} Our method exhibits limitations in handling transparent layers, high-frequency details, and rapid motion.}
    \label{fig:failure_cases}
\end{figure}

We present {\ourmethod}, a native 4D generation framework capable of producing high-quality 4D assets from a single monocular video. At its core lies a structured spacetime latent space, seamlessly extended from a pre-trained 3D prior. By temporally aligning the 3D generator and autoencoder, our method ensures spatio-temporal consistency. To address redundancy across time, we leverage factorized 4D convolutions for effective temporal compression. A tailored training strategy further enhances robustness under real-world conditions. Extensive experiments show that {\ourmethod} significantly outperforms existing baselines. We hope this work contributes to the ongoing progress in 4D generation by introducing a native model learned directly from spacetime data.


\paragraph{Limitation and Future Work} 
While our method achieves strong performance in 4D generation, it still exhibits several limitations. First, it adopts a two-stage pipeline inherited from TRELLIS, which results in less efficient training compared to fully end-to-end approaches. Second, as the model is trained primarily on synthetic data, it struggles to generate photo-realistic appearances when applied to real-world inputs, often producing over-simplified textures. Incorporating real-world video data during training could help bridge this gap and improve generalization.

In addition to these structural limitations, our model also exhibits performance degradation in several practical scenarios, as illustrated in Figure~\ref{fig:failure_cases}:
(a) \textbf{Transparent layers}: The model retains only the outermost voxels and discards internal ones, making it incapable of accurately reconstructing or generating transparent or multi-layered objects. For example, in the crystal ball case, the model produces overly blurred textures due to its inability to capture complex light refraction through multiple transparent surfaces.
(b) \textbf{High-Frequency Details}: Despite employing temporal alignment, maintaining consistency in high-frequency details remains challenging. For instance, our model fails to preserve the fine patterns in camouflage pants, resulting in noticeable flickering. Introducing pixel-space losses may help retain finer details more effectively.
(c) \textbf{Rapid Movement}: Scenarios involving fast motion and strong motion blur are particularly difficult. In the dog example, the model fails to reconstruct the correct shape under rapid movement.

\paragraph{Ethical Issues} As with other generative models, our approach may inherit ethical and diversity-related biases from Objaverse and ObjaverseXL. Additionally, there are potential risks of misuse, particularly in generating deceptive or misleading content.

\paragraph{Acknowledgement}
This work was supported by National Key R\&D Program of China 2022ZD0161600, Shanghai Artificial Intelligence Laboratory, Hong Kong RGC TRS T41-603/20-R, the Centre for Perceptual and Interactive Intelligence (CPII) Ltd under the Innovation and Technology Commission (ITC)’s InnoHK. Dahua Lin is a PI of CPII under the InnoHK.

\bibliographystyle{ACM-Reference-Format}
\bibliography{sample-bibliography}

\begin{figure*}[t]
    \vspace{5pt}
	\centering
	\includegraphics[width=1.0\linewidth]{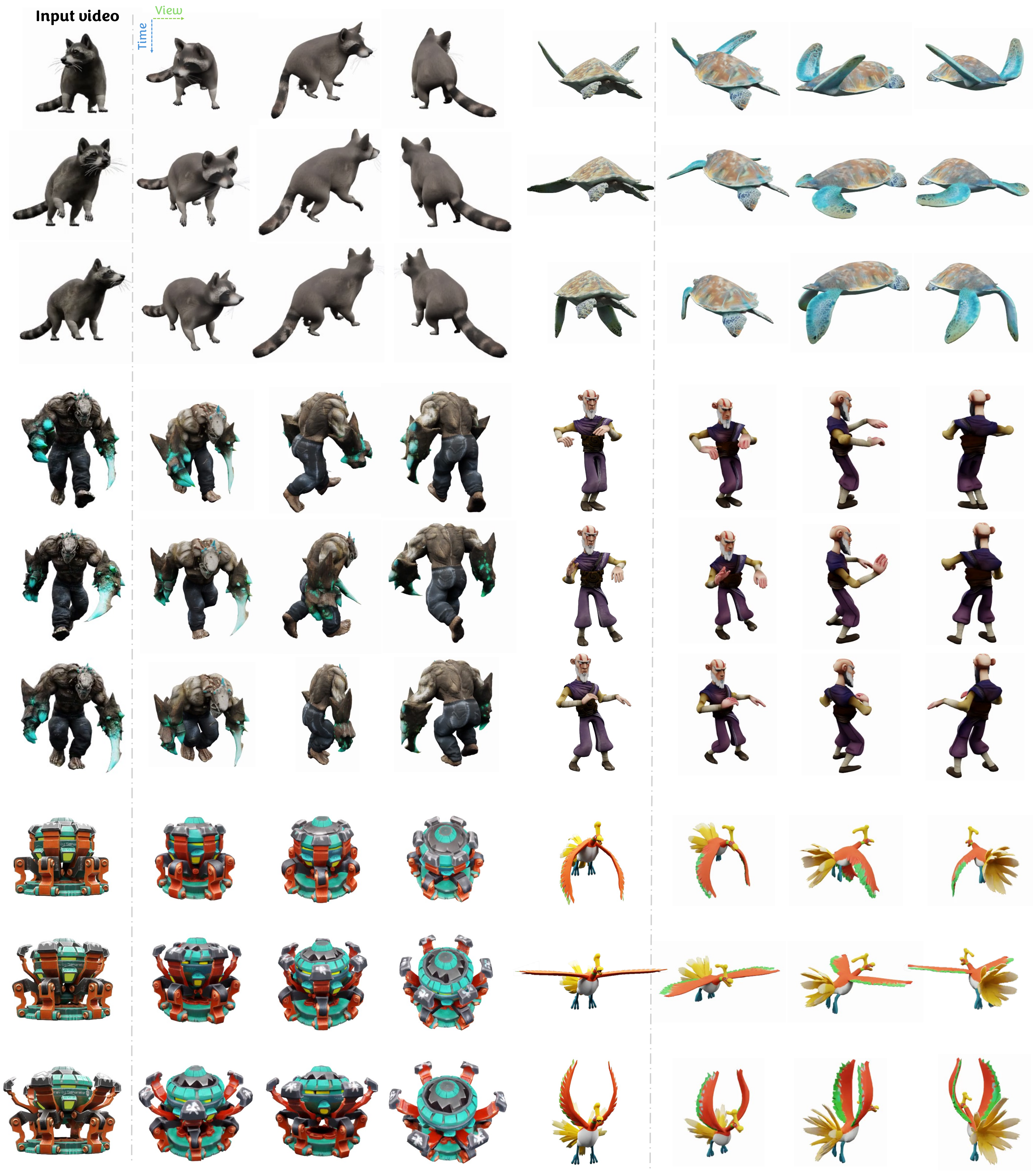}
    \caption{\textbf{Qualitative Results of {\ourmethod} on our Test Set.} Three frames from the input video are shown alongside three corresponding novel view renderings. Our method produces accurate geometry and detailed textures, maintaining consistency over time under significant motion.}
	\label{fig:figure_test}
\end{figure*}

\begin{figure*}[h]
    \vspace{3pt}
	\centering
	\includegraphics[width=1.0\linewidth]{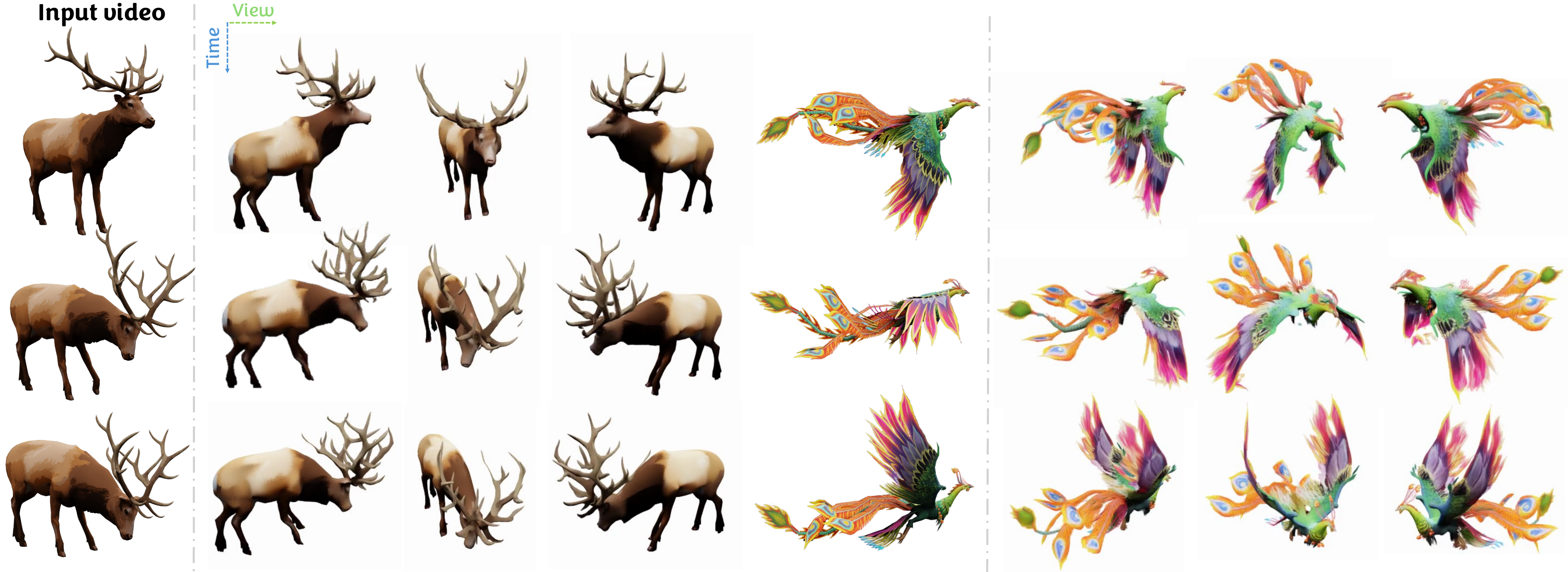}
    \vspace{-15pt}
    \caption{\textbf{Qualitative Results} of {\ourmethod} on \textbf{synthetic data from the Internet}.}
	\label{fig:figure_syn}
\end{figure*}

\vspace{10pt}
\begin{figure*}[h]
	\centering
	\includegraphics[width=1.0\linewidth]{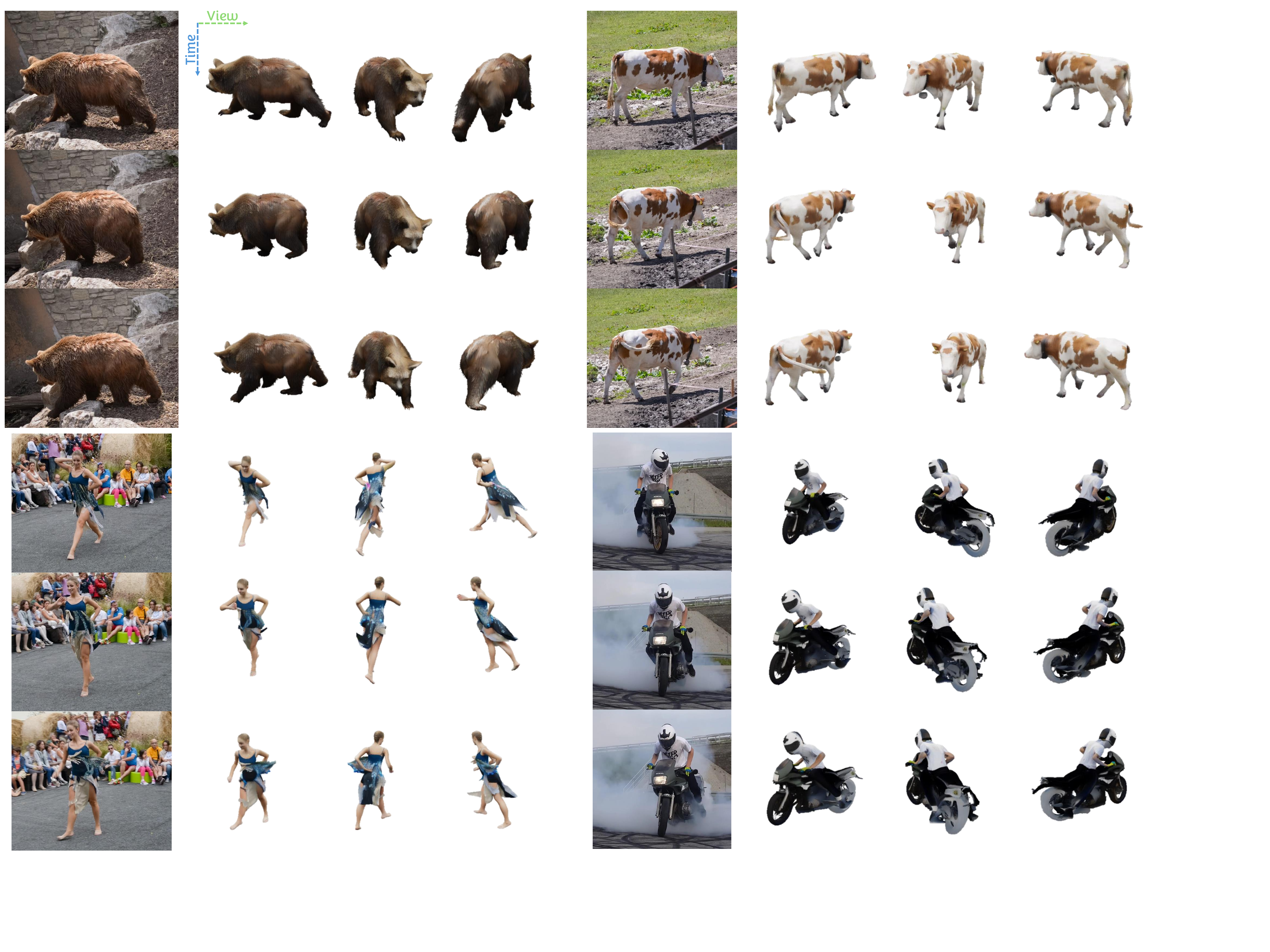}
    \vspace{-15pt}
    \caption{\textbf{Qualitative Results} of {\ourmethod} on \textbf{real-world data}.}
	\label{fig:figure_real}
\end{figure*}

\end{document}